# Unsupervised Learning and Exploration of Reachable Outcome Space

Giuseppe Paolo[1,2], Alban Laflaquière[2], Alexandre Coninx[1] and Stephane Doncieux[1]

*Abstract*— Performing Reinforcement Learning in sparse rewards settings, with very little prior knowledge, is a challenging problem since there is no signal to properly guide the learning process. In such situations, a good search strategy is fundamental. At the same time, not having to adapt the algorithm to every single problem is very desirable. Here we introduce TAXONS, a Task Agnostic eXploration of Outcome spaces through Novelty and Surprise algorithm. Based on a population-based divergent-search approach, it learns a set of diverse policies directly from high-dimensional observations, without any task-specific information. TAXONS builds a repertoire of policies while training an autoencoder on the high-dimensional observation of the final state of the system to build a low-dimensional outcome space. The learned outcome space, combined with the reconstruction error, is used to drive the search for new policies. Results show that TAXONS can find a diverse set of controllers, covering a good part of the ground-truth outcome space, while having no information about such space.

## I. INTRODUCTION

Learning how to control a robot through Reinforcement Learning (RL) in unknown environments is a challenging task, especially in sparse rewards settings. In such situations, a good strategy is to ignore the reward signal and instead to explore the space of possible policies. This approach is used in population-based divergent search algorithms like Novelty Search (NS) [1] or Quality-Diversity (QD) [2], [3], [4] to find as many diverse policies as possible. They work by defining an *outcome space* by hand, and by driving the search for new policies based on a measure of diversity, novelty or surprise in this space. In order for this search to be efficient, the outcome space is designed to be low-dimensional, by selecting a few features that are relevant to characterize the policies.

A benefit of NS and QD approaches is that they are population-based. Instead of looking for a single complex policy able to cover the outcome space, they generate simpler policies, each one specialised in reaching a sub-part of this space. The best policies according to the given metric are then saved into a *repertoire* that is returned as solution to the problem. They have been shown to be useful, for instance, to make a robot resilient to damage [2] or to generate complex behaviours by combining these simple policies in the context of hierarchical RL [3]. Another benefit is that these methods do not use a reward, the search is therefore not mislead by deceiving reward gradients. Furthermore, an outcome space can be shared by different tasks and different domains, meaning that the same repertoire of policies can thus be applied to multiple contexts a posteriori [5], [6], [7], [8].

[1]Sorbonne Université, CNRS, Institut des Systèmes Intelligents et de Robotique, ISIR, F-75005 Paris, France, Email: {alexandre.coninx, stephane.doncieux}@sorbonne-universite.fr
[2]Softbank Robotics Europe, AI Lab, 43 Rue du Colonel Pierre Avia, 75015 Paris, France, Email: {giuseppe.paolo, alaflaquiere}@softbankrobotics.com

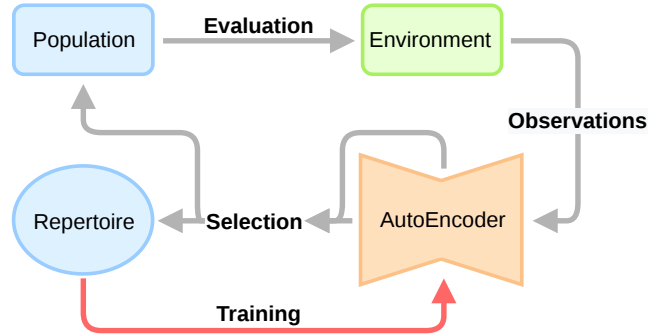

Fig. 1: High level schematic of Task Agnostic eXploration of Outcome space through Novelty and Surprise (TAXONS). It consists of two processes operating in parallel. The first one, the search process (gray arrows), generates a set of new policies, evaluates them in the environment and then stores in the repertoire the best ones according to the selection criteria. The second process, highlighted by the red arrow, is the training of the autoencoder (AE) on the observations collected during the search and evaluation of the policies.

One limitation of divergent search algorithms is the amount of prior knowledge required to design the outcome space. This space needs to be adapted by hand for any new agent and/or environment. Apart from being costly in terms of human resources, defining by hand the appropriate features of the outcome space requires for the experiment designer to know features of the robot, environment and tasks. The search will also be constrained by the biases of the designer's choices.

In this work, we introduce the TAXONS algorithm. It is a method designed to build in parallel, in an unsupervised way, a repertoire of diverse policies and the outcome space in which they are compared. It mixes the exploration dynamics of NS [1] with the representation learning capacity of AEs [9]. NS is a divergent search algorithm that has recently been shown to tend towards a uniform exploration of the outcome space, which is an unbiased strategy in the absence of reward [10]. AEs are a class of neural network architectures commonly used for dimensionality reduction.

A similar method has recently been introduced in [11], where an AE learns an outcome space from raw observations of the internal states of the system. The learned space is then used to ensure diversity among the found policies.

However, using observations of the internal states can be limiting because sometimes such states are not available (e.g. in the case of a robot pushing a ball, the position of the ball might not be known). In contrast, TAXONS learns the search space directly from high-dimensional RGB images of the environment, extracting the features in an unsupervised way. At the same time, the diversity of each policy is assessed not only by using the self-learned outcome space, but also by taking advantage of the information provided by the reconstruction error of the AE. This added measure helps improve both the quality of the search space and the diversity of the generated policies.

In the rest of the paper we will formally describe TAXONS, introduce three experiments on which it will be evaluated, compare it to other baseline methods and discuss the results in detail. The related code is available at: *https://github.com/GPaolo/taxons* .

## II. RELATED WORK

### A. Sparse rewards in RL

Sparse rewards is a well-known problem in RL for which many approaches have been proposed in the literature. Some of them focus on trying to maximize the data efficiency of the policy search [12], [13], [14]. Others avoid deceptive gradients by using an evolutionary approach to improve exploration [15]. Finally, some approaches introduce a task-agnostic exploration phase before exploiting the resulting policy on the task 'to solve [16]. A related but different approach learns an inverse mapping from a hand-designed search space to the policy space in a task agnostic way [17]. This method has been extended in [18] to learn the search space directly from observations (images).

The main limitations of these RL-based approaches is that they learn a single (complex) policy instead of a repertoire of simpler ones. Having a diverse set of policies instead of a single one has a major advantage: if a policy fails, another one can be tried. These failures are not rare, as a learned policy may over-fit the learning conditions and thus be inefficient in new contexts. If learning occurs in simulation with an application in reality, this phenomenon is called the reality gap [19], or the simulation bias [20]. It is one of the main issues with learning in robotics and generating a repertoire of solutions instead of a single one is a way to deal with it [5], [8], [21].

### B. Divergent search methods

Population-based algorithms circumvent two limitations of RL-based approaches. They do not rely on potentially misleading gradients, and generate many simple controllers instead of a complex one. Among those approaches, divergent search algorithms [1], [2], [3], [4], [22], [23] are specifically designed to optimize exploration by generating policies as diverse as possible. The solution that these algorithms provide is a repertoire containing the best policies found during the search process. Divergent search methods have also been extended to take into account the policies' performance on a given task, creating the branch of QD algorithms [24], [25].

One major limitation of all these approaches is that diversity is measured in low-dimensional outcome spaces that are hand-designed, thus requiring more involvement from the system's designer.

### C. Outcome space learning

To circumvent the problem of hand-designing the search space, some representation learning methods have already been combined with population-based approaches in the literature. Some of these methods bootstrap the feature learning either by starting from a general descriptor to learn a more domain-specific one [26], or by combining hand-designed and learned features in a hierarchical way [27].

At the same time, in [28] the authors use a NS-based process to generate space-ships shapes whose shape is encoded in a low-dimensional space using an AE. This AE is trained from scratch on each new generation of shapes, thus limiting the power of the approach by removing any memory of the previous iterations. An AE, combined with MAP-Elites [2], was also used in [29] to generate and classify novel images. However, their AE was trained beforehand on a dataset of images, while TAXONS trains the AE online, during the search process.

As stated in section I, a similar method to TAXONS was introduced in [11], in which an AE learned an outcome space from the observations of the internal states of the system. Such internal states are not always easy to observe or might require a lot of prior knowledge about the system. Compared to this approach, TAXONS works directly on high-dimensional RGB images of the environment while taking advantage not only of the learned space but also of the reconstruction error of the AE, thus circumventing the need to observe internal states and reducing the amount of prior needed to approach the problem.

## III. METHOD

### A. Problem Formulation

In this paper, we use the notation and terminology proposed in [10] and directly inspired from the RL formalism. At the core of NS lies a low-dimensional *behavior space*, or *outcome space*, $\mathcal{B}$ which is used to characterize policies. This space is usually hand-designed and tailored based on prior knowledge about the system and the type of task it might have to fulfill. It is in this space that the novelty of the policies generated by the algorithm is evaluated. Each policy, parametrized by $\theta_i \in \Theta$, is run on the system for $T$ time-steps and generates a trajectory $[s_0, \ldots, s_T]$, where each $s_t$ corresponds to the state of the system at the time-step $t$. This evolution of the system is observed via some sensors, such that they produce a corresponding trajectory of observations $[o_0, \ldots, o_T]$, where $o_t \in \mathcal{O}$ is a potentially under-complete observation of the state of the system at time $t$. An observer function $O_B : \mathcal{O}^T \to \mathcal{B}$ then maps this trajectory of observations to a behaviour description $b_i$, corresponding to a set of hand-designed features. The overall process can be summarized by introducing a behaviour function $\phi$ that maps each policy $\theta_i$ to an outcome description:

$$\phi(\theta_i) = b_i \quad (1)$$

Finally, as in [1], the novelty of a policy $\theta_i$ is defined as the average distance to its $k$ closest previous policies in the outcome space:

$$n(\theta_i) = \frac{1}{k}\sum_{j=1}^{k} dist(\phi(\theta_i), \phi(\theta_j)) \quad (2)$$

The process is applied repeatedly and, at each iteration, new policies are generated and the most novel ones are saved into an archive, ultimately returned as repertoire of diverse policies.

### B. AutoEncoded Novelty and Surprise

As already mentioned, hand-designing the outcome space requires prior knowledge about the robot, the environment and the potential task(s). In situations where it is not clear which features would benefit the search this can hinder the performances of the algorithm.

To overcome these problems, we propose to autonomously build a low-dimensional representation of the observations to be used as behaviour description for novelty estimation. In this work we consider that the last observation, $o_T$, is informative enough to characterize the behaviour of the system. Consequently, only this last observation will be used to build the outcome space. We propose to use an AE's encoder $\mathcal{E}$ as observation function, and its relative feature space $\mathcal{F}$ as outcome space:

$$\begin{aligned}\mathcal{E} : \mathcal{O} \to \mathcal{F} \\ \mathcal{D} : \mathcal{F} \to \mathcal{O}\end{aligned} \quad (3)$$

The AE is trained in an online fashion on the observations generated at each iteration when evaluating the new policies. Note that in this process no task or reward is required.

During the online training, the best policies are selected according to two metrics, ensuring both their novelty and the representativity of the outcome space. The first one, referred to as *novelty*, corresponds to the novelty metric of NS already defined in (2). More precisely, the mapping $\phi$ in (1) is replaced by the mapping:

$$f(\theta_i) = \mathcal{E}\big(o_T^{(\theta_i)}\big), \quad (4)$$

where $o_T^{(\theta_i)}$ is the last observation generated by the policy $\theta_i$.

The second metric, referred to as *surprise*, corresponds to the reconstruction error of the AE; it is expressed as:

$$s(\theta_i) = \big|\big|o_T^{(\theta_i)} - \mathcal{D}(\mathcal{E}(o_T^{(\theta_i)}))\big|\big|^2. \quad (5)$$

This reconstruction error tends to be large when the AE processes observations which have not been frequently encountered yet. By maximising this metric during the training, we ensure that new policies tend to explore novel parts of the state (observation) space. This ensures that the observations are representative of the states the system can reach. In practice, one of the two metrics is picked with a probability of 0.5 to evaluate every new iteration of policies. This strategy is similar to the one used in [30] to mix different behaviour descriptors.

Combining these two metrics drives the search towards an outcome space that is representative of the reachable states of the system and towards policies that are diverse in this space. To our knowledge, TAXONS is the first method to combine these two metrics in such a way.

### C. Search and Training

Similarly to NS, the repertoire of diverse policies is built iteratively. At each iteration, a set of $M$ new policies, parametrized by $\theta \in \Theta$, is generated by modifying the ones from the previous iteration. More precisely, the $Q$ best policies, according to the metric (novelty or surprise), are duplicated to replace the $Q$ worst ones. Then the parameters $\theta$ of all $M$ policies are perturbed by adding gaussian noise with probability $p_d$. Moreover, in the process, the $Q$ best policies $\theta_i$ are also stored in the repertoire, along with their final observation $o_T^{(\theta_i)}$.

The AE is trained to minimize the reconstruction error by feeding it the observations generated during the policies evaluation. In particular, the final observations $o_T$ are stored for $I$ iterations (for a total of $M \times I$ observations) before the AE is trained for $J$ epochs. This buffering step helps in stabilizing the training process of the AE.

Note that, because the outcome space changes during the training of the AE, the policies in the repertoire are reassigned an updated outcome descriptor at each iteration, by feeding the associated final observation to the current version of the AE.

The TAXONS search process is described in algorithm 1.

## IV. EXPERIMENTS

TAXONS was tested in three different simulated environments: a) a two wheeled robot navigating a 2D maze [1], b) a four legged robot moving on the floor [31] and c) a 7−jointed Kuka arm pushing a cube on a table. Each scenario is observed through a top-view RGB-camera for environments a-b) and a side-view RGB-camera for environment c), as illustrated in Fig. 2.(b). We compared TAXONS against four different baselines:

- **NS**: a vanilla novelty search algorithm [1] with hand-crafted features tailored using a priori knowledge about the agent and environment;
- **PNS**: a policy search algorithm, similar to NS but where the outcome space directly corresponds to the parameter space $\Theta$ of the policies. The outcome descriptor characterizes the policy but not the final observation;
- **RNS**: a novelty search algorithm where the outcome description of each policy is randomly sampled in a 10D space. The outcome descriptor does not characterize the observation nor the policy;
- **RS**: a random search in which all policies are randomly generated and randomly selected to be added to the repertoire;

**Algorithm 1:** TAXONS search process.

**Inputs:** population size $M$, environment **env**, training interval $I$, number of best policies $Q$, number of closest policies $k$, episode length $T$, search budget;
**Result:** Repertoire of diverse policies.
initialize autoencoder **ae** and population **pop** with random parameters;
initialize empty Archive **arc**;
initialize empty Buffer **buf**;
**while** *search budget not depleted* **do**
    generate new **pop** of policies $\theta_i$ by random mutation of old **pop**;
    **for** $\theta_i$ *in pop* **do**
        evaluate policies **env**$(\pi(\theta_i)) \rightarrow o_T$;
        calculate outcome descriptor $\mathcal{E}(o_T^{(\theta_i)}) = f(\theta_i)$;
        calculate Surprise as **ae** reconstruction error;
        store outcome observation $o_T \rightarrow$ **buf**;
    **end**
    **for** $\theta_i$ *in pop* **do**
        find $k$ closest policies in outcome space;
        calculate Novelty as
        $n(\theta_i) = \frac{1}{k} \sum_{j=1}^{k} dist(f(\theta_i), f(\theta_j))$;
    **end**
    add Q best policies to **arc**;
    substitute Q worst policies in **pop** with Q best policies;
    **if** *search iteration multiple of I* **then**
        train **ae** $\leftarrow$ **buf**;
        empty **buf**;
        update stored policies' outcome representations;
    **end**
**end**

- **NT**: a novelty search algorithm in which the outcome descriptor is given by the features extracted by a random autoencoder from $o_T$.

The vanilla version of TAXONS is also compared against two ablated versions:

- **TAXO-N**: in which only novelty is used as selection metric;
- **TAXO-S**: in which only surprise is used as selection metric.

In all experiments we used a population with $M = 100$ policies at each iteration. The AE consists in an encoder with 4 convolutional layers, of sizes [32, 128, 128, 64] and 3 fully connected layers, of sizes [1024, 256, 10]; followed by a decoder with 2 fully connected layers, of sizes [256, 512], and 4 convolutional layers of sizes [64, 32, 32, 3]. For the convolutional operations, we used a kernel size of 4 and a stride of 2 with padding of 1. The activation functions used are SeLU [32] for every layer, except for the last layer of the decoder, in which a ReLU activation is used. The training is done every $I = 30$ search iterations for $J = 5$ epochs, with a learning rate of 0.001. The observations $o_T$ consist of RGB images of size $64 \times 64 \times 3$. The novelty of each policy is calculated by using a value of $k = 15$ neighbours in (2), as proposed in [33], with the $Q = 5$ best policies added to the repertoire. Moreover, at each iteration, the parameters $\theta_i$ of each policy are perturbed, with probability $p_d = 0.2$, by adding noise sampled from $\mathcal{N}(0, 0.05)$.

The goal of our approach being to produce diverse policies, we propose to compare the algorithms based on how well they cover the ground-truth outcome space of the system. By design, this ground-truth outcome space corresponds to the $(x, y)$ position of a-b) the center of the robot, or c) the cube. We thus define the *coverage* as the percentage of this $(x, y)$ space reached by the final repertoire of policies. This is done by dividing this space in a $50 \times 50$ grid and then calculating the ratio of number of cells reached at least once over the total number of cells.

Note that the ground-truth $(x, y)$ space is unknown to the methods (except for NS) and is only used a posteriori to compare them.

Moreover, to evaluate the statistical significance of the results, each experiment was run 20 times on different random seeds, and the results compared by performing a Mann-Whitney test [34], with Holm-Bonferroni correction [35].

The evolution of the coverage over the training for the different methods is displayed in Fig. 2.(d) and the final coverage comparison is displayed in Fig. 2.(e).

### A. Maze environment

As illustrated in Fig. 2.(a), the agent consists in a two-wheeled robot, depicted in blue, navigating in a maze, as proposed in [1]. The agent is equipped with 5 distance sensors in the front. The policy controlling the speed of each wheel of the agent is defined by a 2-layers, fully connected, neural network that takes as input the robot sensors readings. The policy is run for a time horizon of 2000 steps. As shown in Fig. 2.(d), the final observation $o_T$ consists in a top-view of the maze and the agent. Note that, for the NS baseline the $(x, y)$ ground-truth position of the robot is used as outcome descriptor.

The search methods are run for 1000 evaluations.

### B. Ant environment

As illustrated in Fig. 2.(a), the agent consists in a four-legged ant robot [36], moving in a 2D plane of size $3m \times 3m$. The policy controlling the torque of each of the 8 agent joints is defined by a sinusoidal Dynamic Movement Primitive (DMP). The experiment is run for a time horizon of 500 steps or until the robot reaches the borders of the plane. As shown in Fig. 2.(d), the final observation $o_T$ consists in a top-view of environment. Note that, for the NS baseline the $(x, y)$ ground-truth position of the robot is used as outcome descriptor.

The search methods are run for 500 evaluations.

### C. Kuka environment

As illustrated in Fig. 2.(a), the agent consists in a 7-jointed Kuka robotic arm simulated in PyBullet [37]. The arm can move thanks to a joint position controller and can push a red cube on a table. Moreover, at the end of the policy execution the arm is reset to the initial position.

As shown in Fig. 2.(b), the final observation $o_T$ consists on a lateral view of the table on which the cube rests. Note that, for the NS baseline the $(x, y)$ ground-truth position of the cube on the table is used as outcome descriptor.

The search methods are run for 1000 evaluations.

## V. Results

The results displayed in Fig. 2.(d-e) show that TAXONS leads to a good coverage of the ground-truth $(x, y)$ outcome space. Its performance is lower than the upper-bound performance of NS, which has direct access to the ground-truth outcome space, but significantly higher than the other baselines, which use a high dimensional outcome space (PNS), a random outcome space (RNS), or no outcome space at all (RS).

This shows that i) performing NS in a low-dimensional outcome space capturing informations about the final state of the system (through the last observation) is beneficial, and ii) that TAXONS successfully builds such a space. Indeed when the generation and selection of policies is purely random (RS) the coverage is very low. Similarly, when low-dimensional outcome descriptors are randomly assigned to the policies the coverage is only slightly better than purely random (Ant), or as bad (Kuka and Maze). At the same time, performing the search on low-dimensional outcome descriptors extracted from the high-dimensional descriptors by a random AE (NT) leads to good performances in the simpler case (Maze) and discrete ones in situation with more complex observations (Ant and Kuka). Finally, performing the NS directly in the high-dimensional policy parameters space $\Theta$ (PNS), leads to a coverage that is similar to the RNS case. This suggests that performing the search in the high-dimensional policy parameter space is equivalent to assigning random descriptors to the policy; the selection process has no information about the actual outcome of the policy. In contrast, the performance of TAXONS is significantly higher and more consistent in the three experiments (Maze: $p = 3.38 \times 10^{-8}$, Ant: $p = 1.6 \times 10^{-7}$, Kuka: $p = 3.94 \times 10^{-8}$). This shows that by learning the outcome space through the AE, it is possible to capture relevant information about each system. The final performance of TAXONS (Maze: 66.02, Ant: 41.55, Kuka: 39.2) even approaches that of NS (Maze: 72.99, Ant: 55.88, Kuka: 63.9) although it remains inferior (Maze: $p = 3.69 \times 10^{-5}$, Ant: $p = 7.65 \times 10^{-8}$, Kuka: $p = 3.39 \times 10^{-8}$). It must be highlighted that NS has direct access to the ground-truth $(x, y)$ space, thus guaranteeing a very good performance.

The performance of the two ablated versions (TAXO-N and TAXO-S) is similar to the vanilla version of TAXONS, as they lay between the NS upper-bound and the PNS, RNS and RS baselines. Nonetheless, their efficiency varies between experiments. TAXO-S performs worse than TAXONS in the Maze ($p = 1.53 \times 10^{-6}$) and in the Kuka environment ($p = 1.34 \times 10^{-6}$) while doing better in the Ant environment ($p = 7.69 \times 10^{-8}$). On the other hand, TAXO-N performs similarly to TAXONS in the Kuka and Maze environments, while being significantly worse in the Ant one ($p = 7.67 \times 10^{-8}$). After investigation, we hypothesize that the low performance of TAXO-N in the Ant environment is due to the specific dynamics of the AE. In the first phase of the training, the AE learns to reconstruct the large body of the agent while disregarding its legs. This leads to the outcome space temporarily capturing informations about the position of the agent in the $(x, y)$ space, and thus allowing novelty search to better cover the ground-truth space. In a second phase, the AE focuses on reconstructing the legs. This stifles the ability of novelty search to cover the $(x, y)$ space, producing a set of policies with different final legs arrangements, rather than final body positions. This second phase does prevent the coverage to improve. TAXO-S performs significantly better in the same environment, as the impact of the body position on the reconstruction error is greater than the one of the legs. Thus maximizing the surprise also leads to maximizing the coverage.

From the results, combining novelty and surprise renders TAXONS more robust to different environments than its two ablated versions, while still being able to perform almost as well as NS.

## VI. Conclusion and Future Work

In this work, we introduced TAXONS, a population-based, task-agnostic exploration algorithm. It can generate a repertoire of diverse policies, without any external reward and with minimal prior knowledge about the system. It does so by applying NS in a low-dimensional outcome space learned online using an AE trained directly on RGB images observations collected during the search.

We tested the approach on three different simulated environments. The results show that, by maximizing both novelty in the learned outcome space and surprise, derived from the AE's reconstruction error, TAXONS finds a set of policies that covers the ground-truth outcome space, while being robust to different environments.

Moreover, even if this feature has not been explicitly shown in the present paper, once the search is over, the learned AE can a posteriori be used to select the policies according to a desired outcome (task) [7], [8]. This can be done by feeding the AE's encoder $\mathcal{E}$ an observation $o_g$ of the desired final state, to extract an outcome descriptor. The policy with the closest outcome descriptor can then be selected as a solution to the task.

A major limitation of the current method that we plan to overcome in the future is its intrinsic sensitivity to distractors in the environment. This phenomenon can already be seen in the Ant environment, in which the configuration of the legs of the robot, despite being irrelevant to the coverage metric, disrupted the exploration of the ground-truth outcome space.

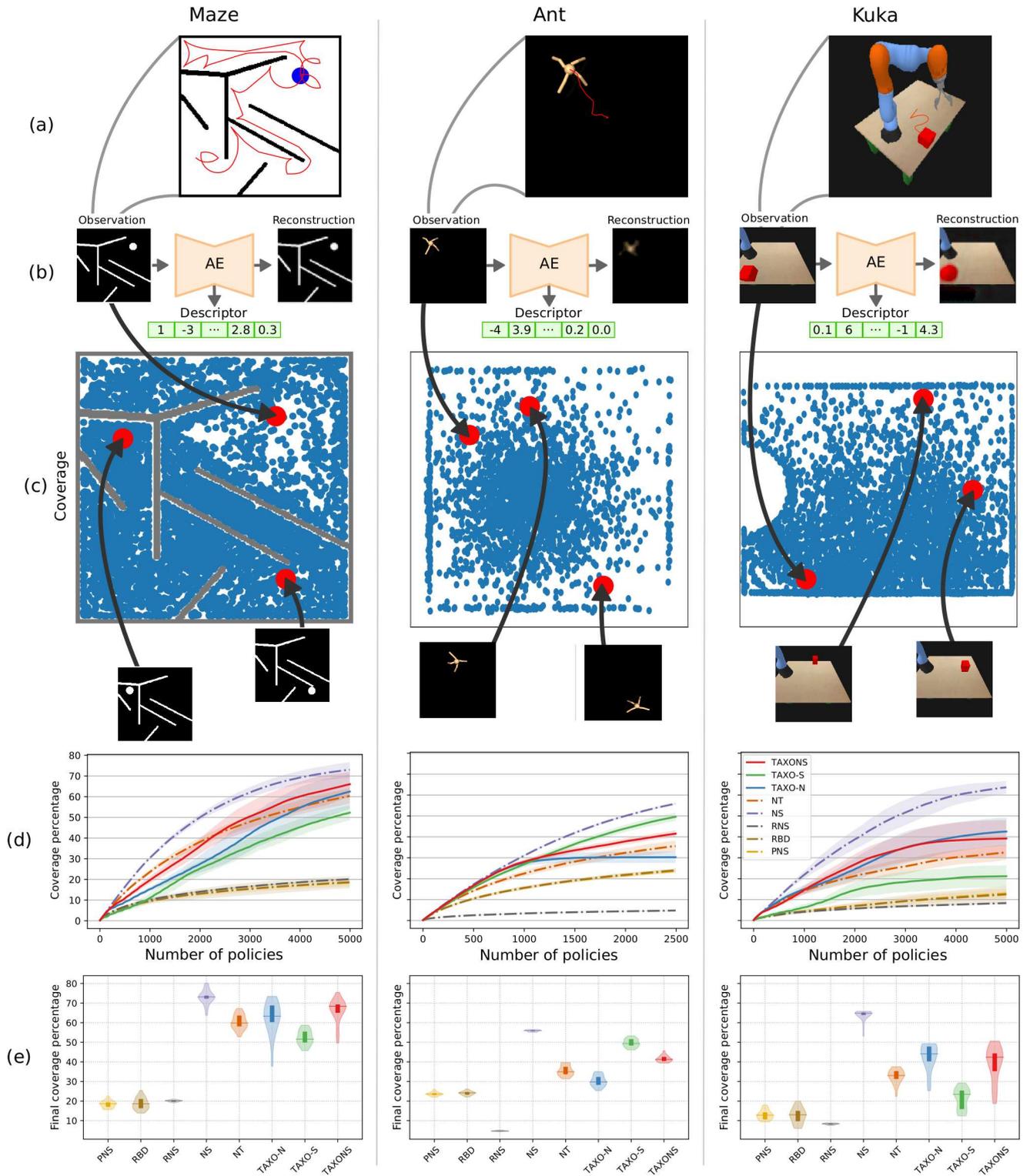

Fig. 2: (a) Sample policy from the repertoire generated by TAXONS. For the Maze and Ant environments, the trajectory of the robot is highlighted. For the Kuka environment, the trajectory of the box is shown. (b) Final observation $o_T$ obtained from the evaluation of the policy shown in (a), with the reconstruction and the outcome descriptor generated by the AE. (c) Coverage of the repertoire of policies, generated by TAXONS, in the ground-truth $(x, y)$ space. Highlighted in red are 3 policies for which the related final observations $o_T$ are shown. (d) Evolution of coverage metric over the number of policies in the repertoire. Note that the PNS and RNS curves are overlapping. (e) Final coverage measure for the tested methods.